\documentclass[10pt,twocolumn]{ICCAS}
 
\usepackage{diagbox}
\usepackage{multirow}
\usepackage{url}  
\usepackage{booktabs}

\begin{document}

\title{Situation Aware Frontier Prioritization for Quadruped Search and Rescue}

\author{Kevin Farias${}^{1}$, Santiago Martin${}^{1}$, Bárbara Flores${}^{1}$, Vinicio Melgar${}^{1}$, Igor Nunes${}^{1}$, \\ Hiago Sodre${}^{1}$, Pablo Moraes${}^{1}$ and Ricardo B. Grando${}^{1*}$}

\affils{ ${}^{1}$Robotics and AI Lab, Technological University of Uruguay, \\
Rivera, Uruguay (ricardo.bedin@utec.edu.uy){\small${}^{*}$ Corresponding author}}

% \thanks{ \noindent
%   This paper is supported by my funding agencies.
%  }

\abstract{Quadruped robots are a promising platform for search and rescue missions because they can navigate cluttered indoor environments that may be restrictive for wheeled systems. However, in unknown rescue scenarios, autonomous exploration must balance map expansion with the likelihood of finding victims, which is not explicitly addressed by classical frontier selection strategies. This paper presents a situation aware frontier prioritization method for single robot quadruped search and rescue. The proposed approach preserves the frontier exploration framework, but extends frontier ranking with information gain, observation deficit, rescue relevance, terrain penalty, and travel cost. The method is evaluated in Gazebo simulation with a quadruped robot in two indoor rescue scenarios with different levels of difficulty. The first scenario is used as a sanity check, while the second introduces stronger clutter and frontier ambiguity. Experimental results show that all methods perform reliably in a simple scenario, whereas in a complex scenario is different. In that setting, the proposed method achieves the highest completion rate and the highest victim recovery among the evaluated approaches. These results indicate that situation aware frontier prioritization is beneficial when frontier choice becomes nontrivial and rescue utility must be balanced against generic exploration objectives.}


\maketitle

% %-----------------------------------------------------------------------

\section{Introduction}

Search and rescue missions require robotic systems that can explore unknown environments, maintain progress under limited time budgets, and improve the chance of finding victims. In indoor disaster settings, the exploration problem is not limited to map coverage. Candidate frontiers may differ in accessibility, visibility, and expected rescue value, even when they are all reachable from the current robot pose. As a result, a frontier selection policy that is appropriate for generic exploration may not be the most suitable policy for search and rescue \cite{yamauchi1997,bourgault2002}.

Quadruped robots are a relevant platform for this problem because they can move through cluttered spaces and maintain locomotion in conditions that may be restrictive for wheeled systems \cite{raibert2008}. This mobility makes them attractive for reconnaissance and victim search in partially structured environments. However, the advantage of legged mobility alone is not sufficient. Once several exploration alternatives become available, the robot must decide not only where new space can be observed, but also where rescue relevant progress is more likely to be achieved. This difficulty becomes more evident in environments where the layout contains multiple branches, partial occlusions, and clutter that weakens the usefulness of purely geometric frontier selection.

Classical nearest frontier exploration remains a strong baseline because of its simplicity and low computational cost \cite{yamauchi1997}. Information gain exploration improves upon this idea by preferring actions that are more useful for map expansion \cite{bourgault2002}. Risk aware exploration introduces an additional penalty for unsafe or uncertain regions and may improve robustness in cluttered conditions \cite{ott2022}. Although these strategies are effective for autonomous exploration, they are not explicitly designed for rescue oriented frontier selection. In search and rescue, the frontier that is most informative or nearest may not be the frontier that best supports victim discovery.

\begin{figure*}[t]
    \centering
    \includegraphics[width=\textwidth]{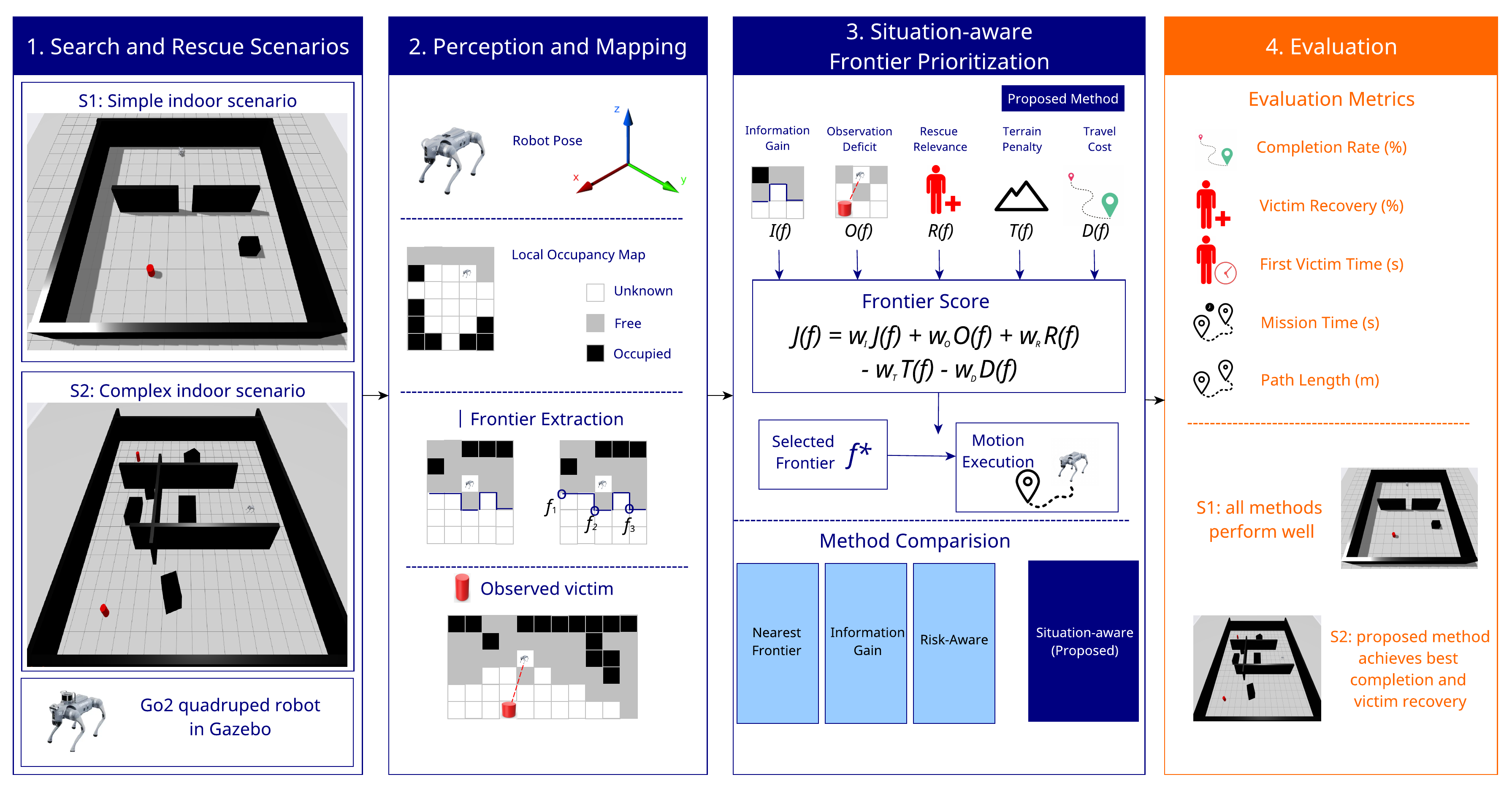}
    \caption{Overview of the proposed framework. The figure summarizes the two rescue scenarios used in the evaluation, the perception and mapping pipeline, the situation aware frontier prioritization strategy, and the main evaluation metrics.}
    \label{fig:overview}
\end{figure*}

In this work, a situation aware frontier prioritization strategy is proposed for single robot quadruped search and rescue. The method preserves the frontier exploration structure, but extends frontier scoring with rescue relevance, observation deficit, terrain penalty, and travel cost. In this way, map expansion is maintained while preference is given to frontiers that are more likely to improve rescue performance. An overview of the proposed framework, including the two evaluation scenarios, the perception and mapping pipeline, the frontier prioritization mechanism, and the main evaluation criteria, is presented in Fig.~\ref{fig:overview}.

The method is evaluated in two simulated indoor environments. The first scenario is intentionally simple and is used as a sanity check. The second scenario introduces a higher level of clutter and branching structure, and is used as the main comparison case. The results indicate that the easy scenario is solved by all methods, while the larger and more complex scenario is different. In that setting, the proposed method achieves the highest completion rate and the highest victim recovery. These results suggest that situation aware frontier prioritization is most beneficial when frontier choice is no longer trivial and rescue utility must be balanced against coverage.

The main contributions of this work are summarized as follows:
\begin{itemize}
    \item A situation aware frontier prioritization method for single robot quadruped search and rescue, in which frontier selection is guided by information gain, observation deficit, rescue relevance, terrain penalty, and travel cost.
    \item A simulation based evaluation protocol for quadruped search and rescue in two indoor environments with different levels of difficulty, including a simple scenario and a complex scenario designed to reveal differences in frontier selection behavior.
    \item An experimental comparison against nearest frontier, information gain, and risk aware baselines, showing that the proposed method achieves the best completion rate and victim recovery in the complex scenario.
\end{itemize}

The remainder of this paper is organized as follows. Section II reviews the related literature. Section III describes the proposed methodology. Section IV presents the experimental results and discussion. Section V concludes the paper and outlines future validation. To support reproducibility, the code and simulation setups are provided in an open-source repository: \url{https://github.com/ricardoGrando/go2_rescue_eval}. A supplementary video can be found at: \url{https://www.youtube.com/watch?v=BbtPfF-NLac}
\section{Related Work}

Frontier based exploration is one of the most established approaches for autonomous exploration in unknown environments. In the classical formulation, a frontier is defined as the boundary between known free space and unknown space, and the robot is guided toward one of these boundaries to expand the mapped region \cite{yamauchi1997}. This approach is attractive because of its simplicity, but it does not by itself distinguish between frontiers that are equally reachable yet different in task relevance.

Information based exploration extends this idea by ranking candidate frontiers according to their expected utility for mapping, commonly through entropy reduction or information gain \cite{bourgault2002}. This formulation is stronger than pure nearest frontier selection because it attempts to maximize map improvement rather than simply minimizing travel distance. However, its objective is still primarily exploratory. In search and rescue, information gain alone may not be sufficient if the most informative frontier is not the most relevant for victim discovery.

Risk aware exploration introduces a different emphasis by accounting for the probability of unsuccessful execution, terrain difficulty, or uncertainty during exploration \cite{ott2022}. These methods are valuable in challenging environments because they reduce the chance of aggressive but unsafe behavior. However, risk aware methods are still not necessarily rescue aware. A frontier may be safe and informative while still being weakly connected to the victim search objective.

In the search and rescue domain, robotic systems are expected to combine locomotion, reconnaissance, and perception under uncertainty \cite{schneider2019,habibian2020}. This motivates task aware exploration strategies that go beyond pure coverage. In the case of quadrupeds, the benefit of legged mobility is not only obstacle negotiation, but also the ability to operate in clutter where the consequences of a poor frontier choice are more pronounced \cite{raibert2008}. The proposed methodology differs from nearest frontier selection by not using travel distance as the primary criterion, differs from information gain exploration by not optimizing map utility alone, and differs from risk aware exploration by explicitly incorporating rescue relevance into frontier ranking. The method is therefore positioned as a rescue oriented frontier strategy rather than as a general mapping strategy.

Although frontier based exploration, information driven exploration, and risk aware navigation provide strong foundations for autonomous search, an important gap remains for quadruped search and rescue. Existing methods are primarily designed for generic exploration objectives such as coverage, uncertainty reduction, or safe motion, but they do not explicitly rank frontiers according to rescue utility. In search and rescue, this is a relevant limitation, because a frontier that is nearest or most informative for mapping may not be the frontier that best supports victim discovery. This gap becomes more pronounced in cluttered environments where frontier choice is no longer trivial.

This work differs from prior nearest frontier, information gain, and risk aware exploration methods by introducing a rescue oriented frontier selection strategy for a quadruped robot. The proposed method preserves the frontier exploration framework, but augments it with rescue relevance and observation deficit in addition to terrain penalty and travel cost. As a result, the method is designed not only to expand the map, but also to bias exploration toward frontiers that are more likely to improve rescue outcomes. This paper therefore addresses the gap between generic exploration policies and task aware search and rescue exploration, and evaluates that difference in indoor environments with increasing decision complexity.
\vspace{-5mm}
\section{Methodology}

\subsection{Problem formulation}

A single quadruped robot is considered in an indoor search and rescue setting represented as an initially unknown environment. The robot must autonomously explore the map, navigate through cluttered passages, and maximize the probability of finding victims within a fixed mission time budget. Let $f$ denote a candidate frontier extracted from the current occupancy map. The decision problem is to select, at each planning cycle, the frontier that best balances map expansion and rescue oriented utility.

The proposed method assigns a score to each frontier according to
\begin{equation}
\begin{aligned}
J(f) =\;& w_I I(f) + w_O O(f) + w_R R(f) \\
       & - w_T T(f) - w_D D(f)
\end{aligned}
\label{eq:frontier_score}
\end{equation}
where $I(f)$ is the information gain associated with the frontier, $O(f)$ is an observation deficit term, $R(f)$ is a rescue relevance term, $T(f)$ is a terrain or traversability penalty, and $D(f)$ is a travel cost term. The selected frontier is the one with the highest score.

This formulation preserves the simplicity of frontier based exploration, but extends it toward search and rescue objectives. Information gain promotes continued map expansion, observation deficit favors regions that remain weakly explored, rescue relevance gives preference to frontiers associated with higher victim search potential, terrain penalty discourages difficult or risky regions, and travel cost avoids unnecessarily expensive motion.

\subsection{System architecture}

The complete system is organized as a perception, planning, and execution pipeline implemented in ROS~2 and evaluated in Gazebo simulation with a Unitree Go2 quadruped model. The architecture contains four main modules: mapping, frontier generation and scoring, victim handling, and local motion execution.

\subsubsection{Mapping and state estimation}

The robot uses onboard sensing to incrementally update a local occupancy representation of the environment. This map is used to distinguish known free space, occupied space, and unknown regions. In parallel, the robot state estimation pipeline provides the pose required for map updates, frontier extraction, and motion control. The mapping layer therefore serves as the common information source for all evaluated methods.

\subsubsection{Frontier extraction and scoring}

At each planning cycle, frontier cells are extracted as boundaries between explored free space and unknown space, following the classical frontier exploration principle \cite{yamauchi1997}. Candidate frontiers are clustered and ranked. The proposed method then evaluates each candidate through the score in (\ref{eq:frontier_score}). In practice, this means that the robot does not only ask which frontier is nearest or most informative, but also which frontier is more relevant for rescue progress.

The information gain term estimates the expected utility of a frontier for expanding the current map. The observation deficit term emphasizes regions that remain insufficiently observed. The rescue relevance term increases the score of frontiers that are more consistent with tentative victim cues or with spatial regions that have greater rescue utility. The terrain penalty reduces the score of frontiers in locally difficult regions. The travel cost penalizes frontiers that require longer motion. As a result, the selected frontier is expected to remain useful for exploration while being more aligned with the rescue task than purely geometric alternatives.

\subsubsection{Victim handling and confirmation}

Victim handling is implemented as a lightweight visual confirmation process. Candidate victim observations are obtained from visual cues corresponding to victim proxies in the simulated environment. These observations are not immediately counted as confirmed victims. Instead, tentative detections are accumulated, spatially deduplicated, and promoted to confirmed victims only after repeated evidence is observed. This reduces the chance of counting multiple observations of the same target as distinct victims.

This component is important because the proposed frontier prioritization is not intended to replace perception. Instead, it uses tentative victim evidence to influence frontier ranking while keeping confirmation conservative enough to avoid trivial double counting. In this way, the robot remains an explorer first and a local inspector second.

\subsubsection{Local motion execution and recovery}

Once a frontier is selected, the robot follows a common local execution layer that is shared across all methods. This design choice is important for fairness, since performance differences should primarily reflect frontier selection rather than low level control differences. The local controller drives the robot toward a path target, applies obstacle aware speed modulation, and uses recovery behaviors when the robot becomes trapped in narrow areas or corners. The same obstacle handling and escape behavior is used for all approaches so that the comparison remains centered on decision making at the frontier selection level.

\subsection{Proposed method: situation-aware frontier prioritization}

The proposed method, denoted Situation-aware Frontier Prioritization, augments frontier based exploration with rescue oriented decision variables. Its main idea is that frontier selection should reflect not only geometric or informational value, but also the likelihood that a frontier supports meaningful search and rescue progress.

Given the set of candidate frontiers $\mathcal{F}$ extracted from the current occupancy map, the selected frontier is defined as
\begin{equation}
f^{*}=\arg\max_{f\in\mathcal{F}} J(f),
\label{eq:frontier_selection}
\end{equation}
where $J(f)$ is the frontier score defined in (\ref{eq:frontier_score}).

To make rescue relevance explicit, the term $R(f)$ is modeled as a function of tentative victim evidence:
\begin{equation}
R(f)=\sum_{v\in\mathcal{V}_t} s_v
\exp\!\left(
-\frac{\|p_f-p_v\|^2}{2\sigma_R^2}
\right),
\label{eq:rescue_relevance}
\end{equation}
where $\mathcal{V}_t$ is the set of tentative victim observations, $p_f$ is the frontier position, $p_v$ is the position of tentative victim cue $v$, $s_v$ is the confidence associated with that cue, and $\sigma_R$ controls the spatial influence of rescue relevance. In this way, frontiers that are closer to stronger tentative victim evidence receive higher rescue relevance, while the final ranking remains balanced by information gain, observation deficit, terrain penalty, and travel cost.

Compared with generic exploration, the proposed method differs in three ways. First, it introduces rescue relevance into frontier ranking, which biases the robot toward regions that are more likely to improve victim discovery. Second, it includes observation deficit so that weakly observed regions are revisited when needed instead of being ignored after minimal coverage. Third, it preserves terrain and travel penalties to avoid rescue relevance dominating motion feasibility.

\subsection{Baseline methods}

After defining the proposed method, three baseline strategies are used for comparison. The first baseline is classical nearest frontier exploration \cite{yamauchi1997}. This method selects the reachable frontier with the lowest travel distance relative to the current robot pose. Its main advantage is simplicity and low computational cost. However, it does not explicitly consider whether a frontier is informative, safe, or rescue relevant. The second baseline is information gain exploration \cite{bourgault2002}. In this case, frontier selection is guided by expected map utility rather than only travel distance. This baseline is stronger than nearest frontier because it explicitly attempts to maximize exploration benefit, but it still treats the task primarily as a mapping problem. It does not incorporate rescue relevance directly into the frontier score. The third baseline is risk aware exploration, which introduces a stronger penalty for uncertain or potentially problematic regions \cite{ott2022}. This baseline is particularly relevant in cluttered environments because it encourages more conservative motion and safer exploration choices. However, as with the previous baselines, its objective is not explicitly rescue oriented.

The proposed method differs from these baselines because it does not optimize only for distance, map utility, or safety. Instead, it introduces rescue relevance as an explicit term in frontier ranking while preserving the overall frontier exploration framework. Therefore, the comparison is not between frontier exploration and a different planning paradigm, but between generic frontier selection policies and a rescue aware frontier policy.

\subsection{Experimental setup}

The experiments are performed in Gazebo simulation using a Unitree Go2 quadruped and a ROS~2 based evaluation framework. All methods share the same quadruped model, sensing stack, local controller, and recovery logic. This ensures that the comparison focuses on the exploration strategy rather than on differences in vehicle dynamics or local navigation implementation.

\begin{table*}[t]
\centering
\caption{Performance comparison in the easy scenario (S1) and the complex scenario (S2). Completion rate is reported as successful trials over 20 runs. Victim recovery corresponds to the mean number of confirmed victims per run. First victim time is averaged over runs in which at least one victim was detected. Bold indicates the best result in each metric within the same scenario. For completion rate and victim recovery, higher is better. For first victim time, mission time, and path length, lower is better.}
\label{tab:s1_s2_results}
\resizebox{\textwidth}{!}{%
\begin{tabular}{|c|l|c|c|c|c|c|}
\hline
\textbf{Scenario} & \textbf{Method} & \textbf{Completion Rate} & \textbf{Victim Recovery} & \textbf{First Victim Time (s)} & \textbf{Mission Time (s)} & \textbf{Path Length (m)} \\
\hline
\multirow{4}{*}{S1}
& SA Frontier Prioritization & 15/20 (75\%) & 0.75 & 396.7 & 548.7 & 79.11 \\
\cline{2-7}
& Information Gain & \textbf{19/20 (95\%)} & \textbf{0.95} & \textbf{392.8} & \textbf{424.2} & \textbf{58.67} \\
\cline{2-7}
& Nearest Frontier & 17/20 (85\%) & 0.85 & 410.1 & 499.6 & 70.91 \\
\cline{2-7}
& Risk-aware Exploration & 17/20 (85\%) & 0.85 & 443.6 & 528.2 & 74.62 \\
\hline
\multirow{4}{*}{S2}
& SA Frontier Prioritization & \textbf{20/20 (100\%)} & \textbf{2.00} & 349.4 & \textbf{373.5} & \textbf{57.45} \\
\cline{2-7}
& Information Gain & 14/20 (70\%) & 1.55 & 459.1 & 612.9 & 91.05 \\
\cline{2-7}
& Nearest Frontier & 18/20 (90\%) & 1.80 & \textbf{326.4} & 414.9 & 60.88 \\
\cline{2-7}
& Risk-aware Exploration & 19/20 (95\%) & 1.95 & 343.7 & 391.8 & 60.72 \\
\hline
\end{tabular}%
}
\end{table*}

Two indoor search and rescue scenarios are considered. \textbf{S1} is an easy scenario designed as a sanity check. It contains one victim and limited clutter. The purpose of this environment is to verify that all methods are able to complete a basic rescue exploration task under relatively simple conditions. \textbf{S2} is a complex scenario and serves as the main evaluation environment. It contains two victims, stronger clutter, and a more ambiguous layout with multiple competing frontier choices. This scenario is specifically intended to expose differences in frontier prioritization behavior and to test whether rescue aware selection provides an advantage once the task is no longer trivial.

Each method is evaluated for 20 runs in each scenario. Runs are initialized with randomized starting pose perturbations in order to reduce dependence on a single initial condition, while all missions use the same time budget and the same local control pipeline. Performance is assessed through completion rate, defined as the number of successful runs over the total number of runs, victim recovery, defined as the mean number of confirmed victims per run, first victim time, defined as the average time required to confirm the first victim over runs in which at least one victim is detected, mission time, defined as the total elapsed mission time, and path length, defined as the total traveled distance during the mission.

\section{Results}

Table~\ref{tab:s1_s2_results} summarizes the results obtained in the easy scenario (S1) and the complex scenario (S2). In S1, all methods achieve relatively high completion rates, which confirms that the environment functions as a valid sanity check scenario. In this easier setting, the Information Gain baseline achieves the highest completion rate and victim recovery, while the proposed method remains functional but does not surpass the strongest baseline. This suggests that explicit situation awareness is less relevant when frontier selection is simple and most candidate frontiers are similarly useful.

\begin{figure}[b]
    \centering
    \includegraphics[width=\columnwidth]{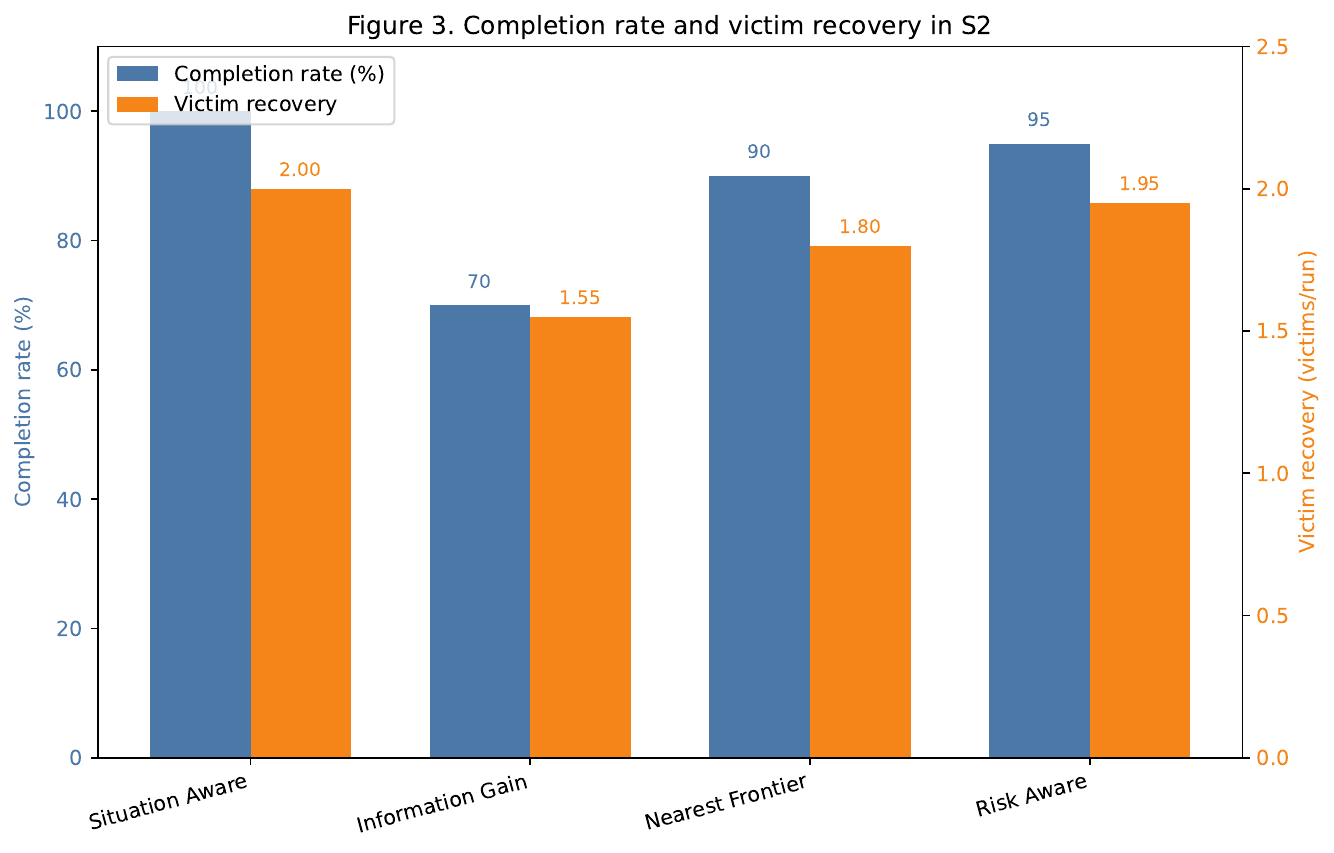}
    \caption{Comparison of completion rate and victim recovery in Scenario S2. The proposed Situation Aware method achieves the highest completion rate, while maintaining competitive victim recovery relative to the baseline methods.}
    \label{fig:s2_completion_recovery}
\end{figure}

A different pattern is observed in S2. In this complex environment, the proposed Situation-aware Frontier Prioritization method achieves the highest completion rate, reaching 20 out of 20 successful runs. It also achieves the highest victim recovery, with a mean of 2.00 confirmed victims per run. By contrast, the Information Gain baseline completes only 14 out of 20 runs and recovers fewer victims on average. Nearest Frontier and Risk-aware Exploration remain competitive, but both are still outperformed by the proposed method in completion rate and victim recovery.

The mission time and path length values support this interpretation. In S2, the proposed method reaches full mission completion without requiring the longest travel distance. The Information Gain baseline travels substantially farther on average, yet still achieves worse completion and victim recovery. This indicates that the proposed method does not simply increase search effort, but instead allocates that effort more effectively. The main benefit of the proposed methodology therefore appears in frontier decision quality rather than in raw motion alone.

Taken together, the results suggest that situation aware frontier prioritization is most beneficial when the environment contains enough clutter and branching structure for frontier choice to become ambiguous. In the easy scenario, simpler exploration criteria are sufficient. In the complex scenario, however, rescue oriented frontier scoring improves mission reliability and victim recovery.

\begin{figure}[b]
    \centering
    \includegraphics[width=\columnwidth]{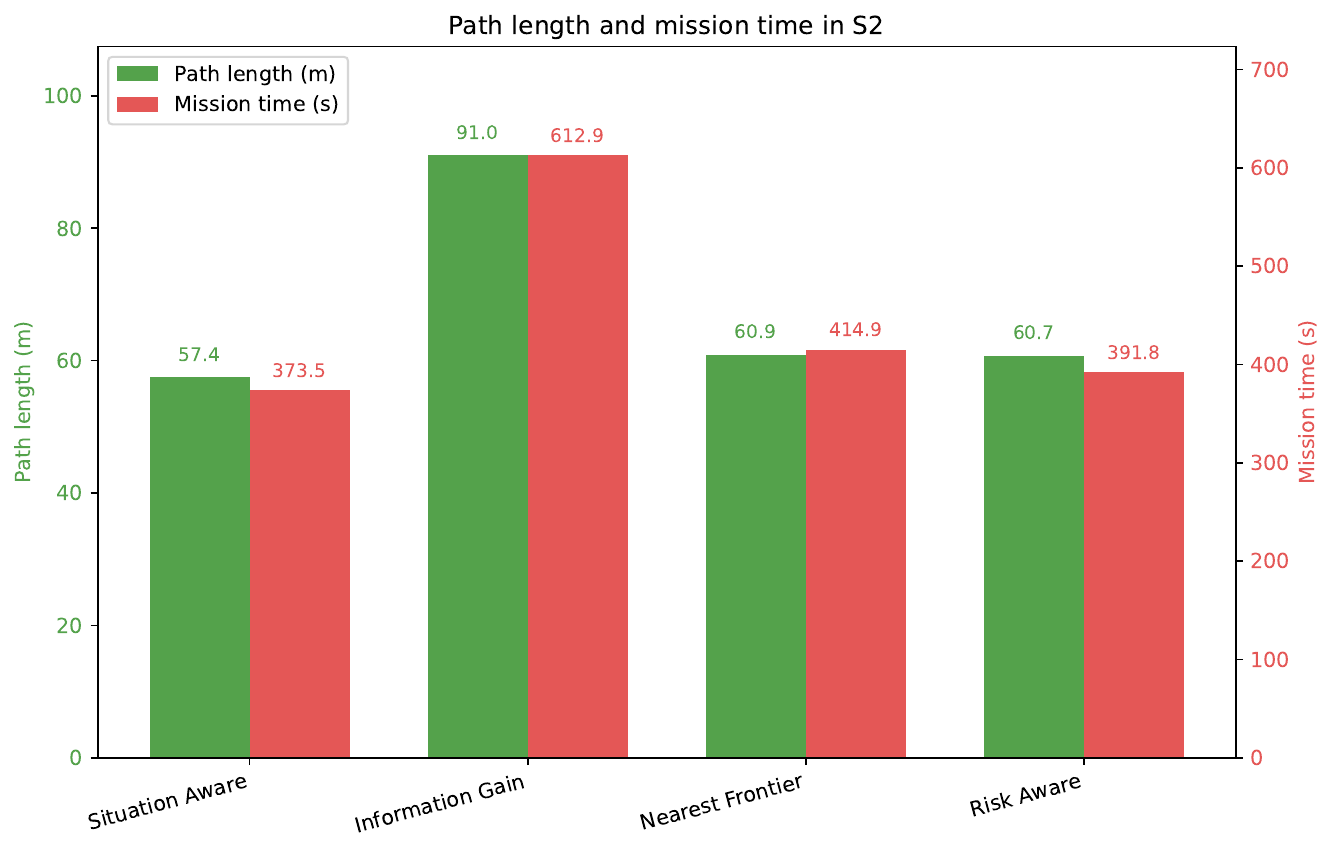}
    \caption{Comparison of path length and mission time in Scenario S2. These results help characterize the exploration behavior of each method in terms of traversal effort and total mission duration.}
    \label{fig:s2_path_time}
\end{figure}

As shown in Fig.~\ref{fig:s2_completion_recovery}, the proposed Situation Aware method achieved the highest completion rate in Scenario S2. In addition, Fig.~\ref{fig:s2_path_time} provides a complementary view of the exploration effort by comparing path length and mission time across methods.

\section{Conclusion}

This paper presented a situation aware frontier prioritization strategy for single robot quadruped search and rescue. The proposed method extends classical frontier exploration with rescue relevance, observation deficit, terrain penalty, and travel cost while preserving the computationally attractive structure of frontier based selection. The method was evaluated in two indoor rescue scenarios representing easy and complex conditions.

The results show that the easy scenario serves as a stable sanity check in which all methods are broadly effective, while the complex scenario is more discriminative. In that setting, the proposed method achieves the highest completion rate and the highest victim recovery, which indicates that rescue aware frontier prioritization becomes useful once frontier choice is no longer trivial. The results therefore support the view that the main value of situation awareness lies in structured cluttered environments where generic exploration objectives are not sufficient by themselves.

Future work should extend the study toward even harder environments, improve victim confirmation under stronger occlusion and uncertainty, and validate the proposed strategy on physical quadruped platforms.

\vspace{-5mm}
\section{Acknowledgements}

The authors of this work would like to thank the Technological University of Uruguay and the Laboratory of Robotics and AI for the support in this work.

\bibliographystyle{./bibliography/IEEEtran}
\bibliography{./bibliography/IEEEabrv,./bibliography/main}

@inproceedings{yamauchi1997,
  author    = {Brian Yamauchi},
  title     = {A Frontier-Based Approach for Autonomous Exploration},
  booktitle = {Proceedings of the 1997 IEEE International Symposium on Computational Intelligence in Robotics and Automation},
  year      = {1997},
  pages     = {146--151}
}

@inproceedings{bourgault2002,
  author    = {Fr{\'e}d{\'e}ric Bourgault and Alexei A. Makarenko and Stefan B. Williams and Ben Grocholsky and Hugh F. Durrant-Whyte},
  title     = {Information Based Adaptive Robotic Exploration},
  booktitle = {Proceedings of the 2002 IEEE/RSJ International Conference on Intelligent Robots and Systems},
  year      = {2002},
  pages     = {540--545}
}

@article{ott2022,
  author  = {Joshua Ott and Sung-Kyun Kim and Amanda Bouman and Oriana Peltzer and Mamoru Sobue and Harrison Delecki and Mykel J. Kochenderfer and Joel Burdick and Ali-akbar Agha-mohammadi},
  title   = {Risk-aware Meta-level Decision Making for Exploration Under Uncertainty},
  journal = {arXiv preprint arXiv:2209.05580},
  year    = {2022}
}

@inproceedings{raibert2008,
  author    = {Marc Raibert and Kevin Blankespoor and Gabriel Nelson and Rob Playter},
  title     = {BigDog, the Rough-Terrain Quadruped Robot},
  booktitle = {Proceedings of the 17th World Congress of the International Federation of Automatic Control},
  year      = {2008},
  pages     = {10822--10825}
}

@article{schneider2019,
  author  = {Frank E. Schneider and Dennis Wildermuth},
  title   = {Assessing the Search and Rescue Domain as an Applied and Realistic Benchmark for Robotic Systems},
  journal = {arXiv preprint arXiv:1912.04693},
  year    = {2019}
}

@article{habibian2020,
  author  = {Soheil Habibian and Mehdi Dadvar and Behzad Peykari and Alireza Hosseini and M. Hossein Salehzadeh and Alireza H. M. Hosseini and Farshid Najafi},
  title   = {Design and Implementation of a Maxi-Sized Mobile Robot (Karo) for Rescue Missions},
  journal = {arXiv preprint arXiv:2008.10396},
  year    = {2020}
}

\end{document}